\def\year{2019}\relax
\documentclass[letterpaper]{article} 
\usepackage{aaai19}  
\usepackage{times}  
\usepackage{helvet}  
\usepackage{courier}  
\usepackage{url}  
\usepackage{graphicx}  
\usepackage{amsmath}
\usepackage{amssymb}
\usepackage{multirow}
\usepackage{booktabs}
\begin{document}
%
\title{Angular Triplet-Center Loss for Multi-view 3D Shape Retrieval}
\author{Zhaoqun Li\textsuperscript{1},\thanks{They contributed equally to this work} Cheng Xu\textsuperscript{1},\footnotemark[1] Biao Leng\textsuperscript{1,2,3}\thanks{Corresponding author}\\
\textsuperscript{1}School of Computer Science and Engineering, Beihang University, Beijing, 100191\\
\textsuperscript{2}Research Institute of Beihang University in Shenzhen, Shenzhen, 518057\\
\textsuperscript{3}Beijing Advanced Innovation Center for Big Data and Brain Computing, Beihang University, Beijing, 100191\\
\{lizhaoqun, cxu, lengbiao\}@buaa.edu.cn
}
\maketitle
\begin{abstract}
How to obtain the desirable representation of a 3D shape, which is discriminative across categories and polymerized within classes, is a significant challenge in 3D shape retrieval. Most existing 3D shape retrieval methods focus on capturing strong discriminative shape representation with softmax loss for the classification task, while the shape feature learning with metric loss is neglected for 3D shape retrieval. In this paper, we address this problem based on the intuition that the cosine distance of shape embeddings should be close enough within the same class and far away across categories. Since most of 3D shape retrieval tasks use cosine distance of shape features for measuring shape similarity, we propose a novel metric loss named angular triplet-center loss, which directly optimizes the cosine distances between the features. It inherits the triplet-center loss property to achieve larger inter-class distance and smaller intra-class distance simultaneously. Unlike previous metric loss utilized in 3D shape retrieval methods, where Euclidean distance is adopted and the margin design is difficult, the proposed method is more convenient to train feature embeddings and more suitable for 3D shape retrieval. Moreover, the angle margin is adopted to replace the cosine margin in order to provide more explicit discriminative constraints on an embedding
space. 
Extensive experimental
results on two popular 3D object retrieval benchmarks, ModelNet40 and ShapeNetCore 55, demonstrate the effectiveness of our proposed loss, and our method has achieved state-of-the-art results on various 3D shape datasets. 
\end{abstract}

\begin{figure*}
\centering
\includegraphics[width=0.9\linewidth]{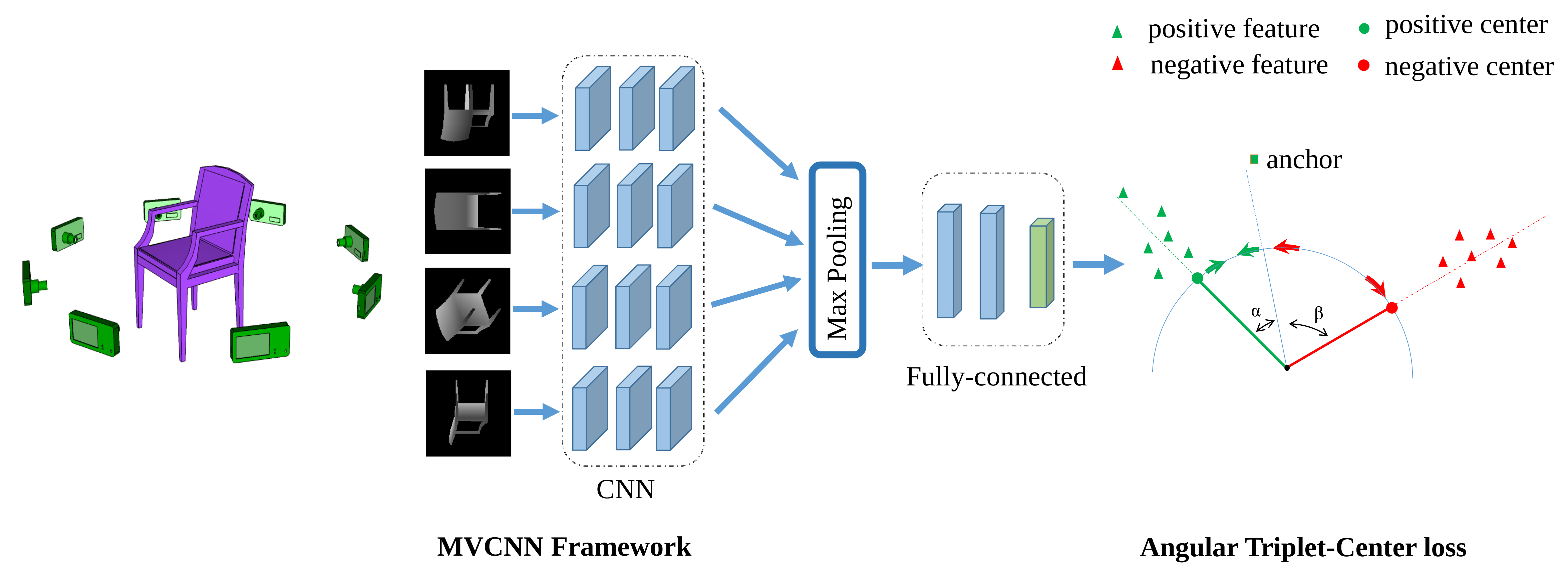}
\caption{The flowchart of the proposed 3D shape retrieval framework. We use MVCNN to extract model features and the proposed ATCL is the supervision signal. 
In the training process, the ATCL encourages the anchor (extracted feature) to minimize the angle with its corresponding positive center while maximizing the angles with other negative centers. }
\label{fig_flowchart}
\end{figure*}

\section{Introduction}
With the development of 3D sensing techniques and availability of large-scale 3D shape repositories~\cite{chang2015shapenet,wu20153d}, 3D shape retrieval has become more significant in 3D shape analysis communities with wide application prospects. Among the existing 3D shape retrieval methods, view-based methods have achieved best performance so far. In view-based 3D shape retrieval, deep shape representation learning over multi-view image sequences of 3D shapes~\cite{xu2018emphasizing,dai2018siamese} have dramatically advanced various 3D shape retrieval tasks in recent years. 

The most fundamental issue in 3D shape retrieval is how to capture the efficient shape embeddings that make the instances from the same category closer and those from different categories farther away. A well-known general retrieval algorithm of view-based methods is Multi-view Convolutional Neural Networks (MVCNN)~\cite{su2015multi},  extracting the visual feature of each view and aggregating them into a compact shape representation in a unified framework. Most view-based methods~\cite{leng2018learning,leng20163d} like MVCNN build discriminative shape features for the object classification task with the softmax loss. However, softmax loss only focuses on distinguishing features across classes and it ignores the intra-class similarity, which is not specifically suitable for 3D shape retrieval task. Recently,~\cite{he2018triplet} firstly formally introduce deep metric learning in 3D shape retrieval community to tackle this problem, arguing that the metric loss such as triplet loss or center loss can bring the performance benefits to 3D shape retrieval. Moreover, they propose the triplet-center loss (TCL) that can optimize the features by minimizing the intra-class variance
while also maximizing the inter-class variance at the same time. Inspired by their work, we develop our method with regard to the deep metric learning in view-based retrieval methods.

Most 3D shape retrieval approaches based on the metric loss like TCL~\cite{he2018triplet} use the Euclidean distance of features to optimize the embedding space in the training, while the cosine distance is adopted for measuring shape similarity in the testing. This inconsistent metric restricts the feature learning of these methods. Moreover, the desirable Euclidean margin is difficult to determine because the range of Euclidean distance between features could be very large.

In this paper, we propose the angular triplet-center loss (ATCL) to build the discriminative and robust shape embeddings for multi-view 3D shape retrieval. Since 3D shape retrieval methods usually adopt cosine distance for measuring feature similarity, the ATCL is designed based on the simple intuition that the cosine distance of features should be close enough within the same class and far away across different classes. Motivated by TCL, the proposed metric loss learns a center for each class and aims to make the cosine distances between an instance and its corresponding center smaller than the distance between it and other centers of different classes. Moreover, considering the metric loss based on cosine margin $\cos(\theta)+m$ suffers from unstable parameter update and is hard to converge if the number of a mini-batch is small, we take an equivalent alternative way that the angular margin $(\theta+m)$ is adopted to directly optimize shape features in angular space based on both L2 normalized instance features and center features. In our method, ATCL is built on the MVCNN framework for view-based 3D shape retrieval that unifies the shape feature learning and deep metric learning into an end-to-end fashion. 
The flowchart of our method is shown in Fig. \ref{fig_flowchart}. 
The rendered images of a shape are input to the MVCNN to generate a shape feature.
Then the feature embeddings are learned under the supervision of our angular triplet-center loss.

Our proposed metric loss can obtain highly discriminative and robust shape representations and has the following key advantages. Firstly, the decision boundary of angular margin has more explicit discriminative constraints on an embedding space than those retrieval methods based on Euclidean distance. Meanwhile, it inherits the property of TCL to minimize the intra-class distance and maximize inter-class distance at the same time. In addition, the proposed loss is easy to converge and the training is stable to update the parameters, especially for large-scale 3D shape dataset like ShapeNetCore55~\cite{chang2015shapenet}. Our method achieves the state-of-the-art performance on ModelNet dataset and ShapeNetCore55 dataset.

In summary, our main contributions are as follows.
\begin{itemize}
\item We investigate the angular metric learning that the shape features are directly optimized in angular embedding space for view-based retrieval, which is more suitable for the nature of 3D shape retrieval task.
\item We propose a novel loss function named angular triplet-center loss that provides more explicit discriminative constraints on an embedding space than those methods based on Euclidean distance.
\item Our method achieves the state-of-the-art on both ModelNet and ShapeNetCore55 datasets, which demonstrates that our shape features are more discriminative.
\end{itemize}

\begin{figure*}
\centering
\includegraphics[width=0.9\linewidth]{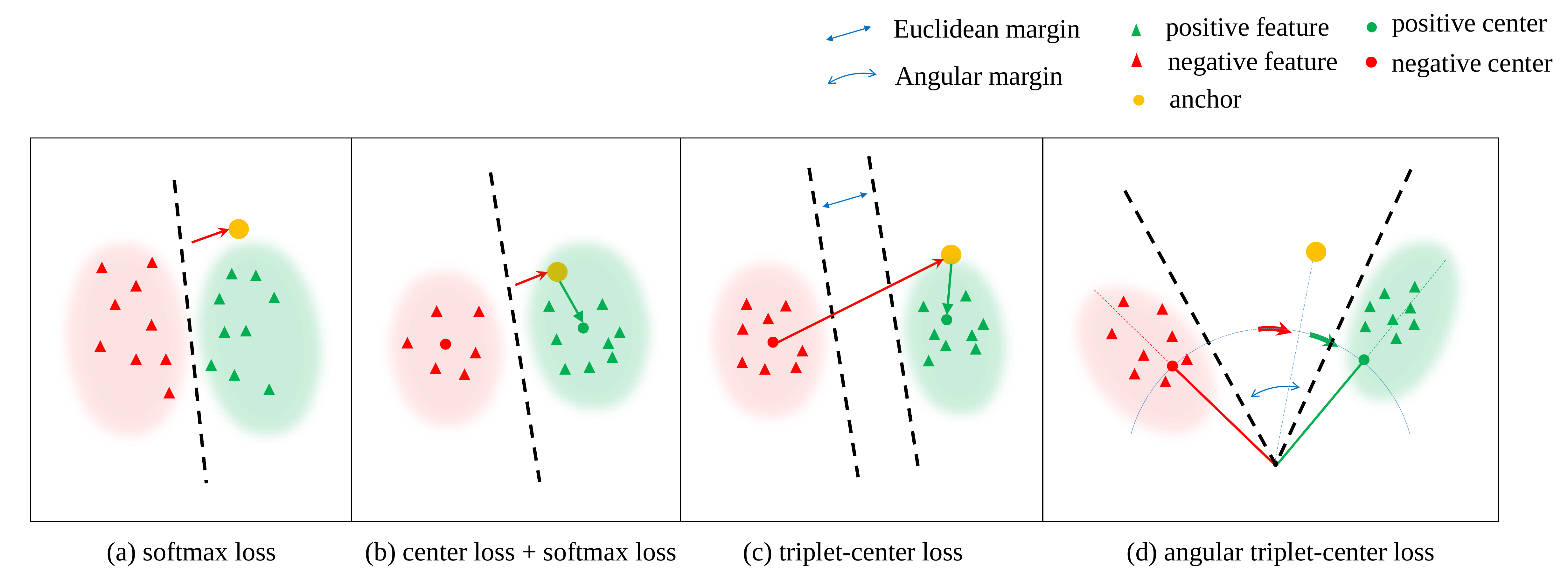}
\caption{The effect of different loss functions on the feature distribution. 
(a) tries to find a decision boundary to separate different classes features, it pushes the anchor away from the decision boundary. 
(b) adds a force to keep the anchor close to its corresponding center. 
(c) pulls the anchor to its corresponding center while it pushes the features away from the other centers
(d) rotates the anchor to its positive center and it augments the angles between the anchor and negative centers.
Compared with (b) (c) which use Euclidean distance, our method conducts the optimization directly in the angular space. }
\label{fig_comploss}
\end{figure*}

\section{Related work}
A large number of works~\cite{ioannidou2017deep} have been proposed to address 3D shape retrieval problem in recent years. In particular, with the explosive growth of large-scale public 3D shape repositories and the success made by the convolutional neural network (CNN) in computer vision, the CNN-based 3D shape retrieval methods have achieved impressive performance. In this section, we will mainly introduce representative 3D shape retrieval methods based on deep learning techniques. 

Generally, 3D shape retrieval methods could be coarsely divided into two categories: model-based methods and view-based methods. The model-based method directly extracts the shape feature from the raw 3D representations of objects, such as polygon meshed~\cite{xie2017deepshape,boscaini2016learning}, voxel grid~\cite{ma2018binary,li2016fpnn}, point cloud~\cite{qi2017pointnet,qi2017pointnet++}, or octree representation~\cite{wang2017cnn}. For example, ~\cite{qi2016volumetric} introduce two volumetric CNN network architectures including the auxiliary part-based classification task and the long anisotropic kernel for long-distance interactions, which improves the performance of volumetric CNN-based 3D shape recognition methods.~\cite{shen2018mining} propose kernel correlation and graph pooling to capture local patterns, leading a robust improvement for PointNet-like methods.~\cite{wang2017cnn} performs 3D CNN on the octree representation of 3D shapes, showing more compact storage and fast computation than existing model-based approaches. Although model-based methods can effectively interpret the geometric characteristics of 3D shapes, their performances are limited by the high computational complexity (\emph{e.g.}, voxel representation) and noise of naive 3D shape representation, such as incompleteness or occlusions.

The view-based 3D shape retrieval methods represent a 3D shape with a collection of 2D projections. A typical deep learning example of the view-based technique is GIFT~\cite{bai2016gift}, extracting each single view feature by using CNN with GPU acceleration and adopting the inverted file to reduce computation in distance metrics. Apart from a single view, many researchers focus on building a compact shape feature through an informative multi-view image sequence of the 3D shape. For example,~\cite{su2015multi} use a set of CNN to extract each view's feature and then aggregate information from multiple views into a single and compact shape representation with the element-wise maximum operation.~\cite{huang2018learning} develop a local MVCNN shape descriptors, which generates a local descriptor for any point on the shape and can be directly applicable to a wide range of shape analysis tasks. Recently,~\cite{feng2018gvcnn} propose a view-group-shape framework (GVCNN) to exploit intrinsic hierarchical correlation and discriminability among views, achieving a significant performance gain on the 3D shape retrieval task. Typically, the deep representation of multiple views performs more discriminative for 3D shapes and leads the best retrieval results on various 3D shape datasets.

The above 3D shape retrieval methods are trained to learn the discriminative 3D shape representation under the supervision of softmax loss for the object classification task. However, this classification task is not specifically desirable for 3D shape retrieval task. Inspired the widely adopted deep metric learning in image retrieval / person re-identification task,~\cite{he2018triplet} recently introduce two popular metric learning loss, triplet loss and center loss, in 3D shape retrieval community and achieve state-of-the-art results on ModelNet and ShapeNetCore55 datasets. Triplet loss~\cite{schroff2015facenet} is proposed to separate shape features within one class from those features from different classes by a distance margin. And center loss~\cite{wen2016discriminative} learns the discriminative features via optimizing a center vector for each class and penalizing the distance between the features and their corresponding class centers. Combining the advantages of triplet loss and center loss, TCL is proposed for 3D shape retrieval and it can optimize the features directly by minimizing the intra-class variance
while also maximizing the inter-class variance at the same
time. However, these methods prefer to use the metric loss based on Euclidean distance, while most 3D shape retrieval tasks adopt cosine distance to measure shape similarity. To make this gap, we propose a novel loss called Angular triplet-center loss. In addition,~\cite{wang2018additive} propose the CosineFace loss for face verification, which is similar to our method. However, our loss directly optimizes the angular distance between different features of instances, while CosineFace is built upon the normalized weights and features.

\section{Proposed method}
The goal of our method is to obtain the discriminative shape representation via directly optimizing the geometric distribution of features in cosine (angle) embedding space. 

For the 3D shape retrieval task, softmax loss is often used to guide feature learning over CNN-based architecture. However, softmax loss tends to maximize the inter-class variance, while the constraint of intra-class variance is neglected. In order to improve the retrieval performance, the metric loss functions, such as triplet loss~\cite{schroff2015facenet}, centerloss~\cite{wen2016discriminative} and TCL~\cite{he2018triplet}, are introduced into 3D shape retrieval framework, as illustrated in Fig. \ref{fig_comploss}. These methods separate sample of different classed by the Euclidean decision boundary, while most 3D shape retrieval methods adopt the cosine distance to measure shape similarity. In order to make up this metric gap, we design our angular triplet-center loss, where a more explicitly discriminative constrain of angle decision boundary is utilized to optimize the embedding space. In the next section, we first review on TCL and then propose our loss function.

\subsection{Review on Triplet-Center Loss}
TCL is proposed for 3D model retrieval that leverages the advantage of triplet loss and center loss.
Given a training dataset $ \mathcal{X} \in \mathbb{R}^D $, let $ \mathcal{Y}=\{ y_1,y_2,...,y_{| \mathcal{X} |} \} $ denote the corresponding label set,
$y_i \in \{1,2,...,K\} $ where $K$ is the number of classes.
Deep metric learning aims to train a neural network, denoted by $f_{\theta}(\cdot)$, 
which maps the data in $\mathcal{X}$ onto a new feature space $F \in \mathbb{R}^n $. 
Let $D(\cdot)$ be a metric function in $F$ which can measure the distances between features.

For writing convenience, we use $f_i \in \mathbb{R}^n$ to represent $f_{\theta}(x_i)$ which is the extracted feature from CNN.
Given a batch of training data with $M$ samples, the TCL can be expressed as:

\begin{equation}
    L_{TC} = \sum_{i=1}^{M} \max(D(f_i,c_{y_i}) +m -\min_{j \neq y_i}D(f_i,c_j), 0) 
    \label{equation_tcl}
\end{equation}
where $D(\cdot)$ represents the squared Euclidean distance function:

\begin{equation}
  D(f_i,c_{y_i}) = \frac{1}{2} ||f_i - c_{y_i}||^2 
\end{equation}

Here, $m$ is a manually designed Euclidean margin and
$c_i \in \mathbb{R}^n $ is the center of class $i$.

The TCL mainly contains two tasks: reduce the distance between feature and corresponding center and enlarge the distance between features and negative center. It can effectively minimize the intra-class variance while also maximize the inter-class variance at the same time. However, similar to center loss and triplet loss, TCL still uses Euclidean distance for 3D shape retrieval task which means that the optimization is not direct for cosine distance.  Moreover, the design of margin $m$ is important for a good performance,  while the squared Euclidean margin is not easy to design.

\subsection{Angular Triplet-Center Loss}
\textbf{Motivation.}
To address the above issues, we propose a metric loss called angular triplet-center loss (ATCL) 
that directly optimizes the angular distances between the features.
Intuitively, improving the cosine distance distribution is a more direct way to enhance the discriminative power of the features.
Meanwhile, the angular margin has a very clear geometric interpretation, it corresponds to the angular distance on the hypersphere. 
And the margin design for angular distance is much more simple since the angle is limited in $[0,\pi]$. 
Under these considerations, we design the angular triplet-center loss for the sake of discriminative across categories and polymerized within on class shape representation.

\noindent \textbf{Proposed loss function.}
We assign each class a center $c_i \in \mathbb{R}^n$.
Different from the explanation in center loss and TCL, the center here represents the \textbf{direction} of the corresponding class features.
For writing convenience, we denote $\tilde{v}$ as the normalized vector of $v$ : $\tilde{v} = \frac{v}{||v||_2}$. 
Let $\mathcal{C} = \{c_1 ,c_2 ,...,c_K \} $ denote center set and $c_{hard_i} \in \mathcal{C} \backslash \{ c_{y_i} \} $ be the \emph{nearest negative center} 
which is the closest center to $f_i$ under a metric function $D(\cdot)$.
Here, $hard_i \in \{1, 2,...,K \} $ is the index of the nearest negative center. 
So we have:
\begin{equation}
\min_{j \neq y_i}D(f_i,c_j) = D(f_i, c_{hard_i})
\end{equation}
We call $\alpha_i = \arccos(\tilde{f_i}^\mathrm{T} \cdot \tilde{c}_{y_i})$ \emph{positive angle} to indicate the angle between feature and corresponding center. 
And $\beta_i = \arccos(\tilde{f_i}^\mathrm{T} \cdot  \tilde{c}_{hard_i})$ is called \emph{negative angle}.
Inspired by TCL, we can get a loss function about cosine distance via replacing directly the $D(\cdot)$ in Eq. \ref{equation_tcl} by cosine distance:

\begin{equation}
    \begin{aligned}
    L_{cos} 
    &= \sum_{i=1}^{M} \max(cosd(f_i,c_{y_i}) +m -\min_{j \neq y_i}cosd(f_i,c_j), 0) \\
    &= \sum_{i=1}^{M} \max(\cos(\beta_i) +m - \cos(\alpha_i), 0) 
    \end{aligned}
    \label{equation_cos}
\end{equation}

where 
\begin{equation}
    cosd(f_i,c_j) = 1 - \frac{f_i^\mathrm{T} \cdot c_j}{||f_i||_2\cdot ||c_j||_2} = 1 - \cos(f_i,c_j)
    \label{equation_cosd}
\end{equation}

But in practice,  this form of loss is not easy to optimize.
In the process of implementation, we observe that the drop of loss is not very stable when the cos margin is large ($>\cos\frac{2}{3}\pi$).
Notice that the cosine value has a one-to-one mapping from the cosine space to the angular space when the angle is in $[0,\pi]$.
Therefore, equivalently, we can choose directly the angle between two vectors as their distance. 
Moreover, the angular margin has a more clear geometric interpretation than cosine margin.
With this idea, we propose ATCL by a new metric function:

\begin{equation}
    \begin{aligned}
    L_{ATC} 
    &= \sum_{i=1}^{M} \max(\arccos(\tilde{f_i}^\mathrm{T} \cdot \tilde{c}_{y_i}) +m \\
    & \quad   -\min_{j \neq y_i}\arccos(\tilde{f_i}^\mathrm{T} \cdot \tilde{c}_j), 0) \\
    &= \sum_{i=1}^{M} \max(\alpha_i +m - \beta_i, 0) 
    \end{aligned}
    \label{equation_atcl}
\end{equation}

As shown above, the positive angle and the negative angle are used directly in the loss design. 
With this final version of ATCL, the training can converge with a large scale of angular margin. 

\noindent \textbf{Back-propagation.}
As the Eq. \ref{equation_cos} and \ref{equation_atcl} show, 
the features need to pass a $L_2$-normalization layer to compute the cosine value or angle.
Besides, centers are also normalized in every iteration. 
Here, we first formulate the gradients of normalized features and then that of normalized centers. 

Given a mini-batch with $M$ samples, $L_{ATC}$ can be regarded as the summation of each sample's loss $L_i$:

\begin{equation}
    L_i = \max(\alpha_i +m - \beta_i, 0) 
    \label{equation_atcl2}
\end{equation}

For convenience, we use $ \delta(condition) = 1 $ to indicate $condition$ is true and $ \delta(condition) = 0 $ otherwise. 
Then we have:
\begin{equation}
    \begin{aligned}
    \frac{\partial {L_{ATC}}}{\partial \tilde{f_i}}
    &= \frac{\partial {L_i}}{\partial \tilde{f_i}} \\
    &= \frac{\tilde{c}_{hard_i} \cdot \delta(L_i>0)}{ \sqrt{ 1-(\tilde{f_i}^\mathrm{T} \cdot \tilde{c}_{hard_i})^2 } }  \\
    &\quad - \frac{\tilde{c}_{y_i} \cdot \delta(L_i>0)}{ \sqrt{ 1-(\tilde{f_i}^\mathrm{T} \cdot \tilde{c}_{y_i})^2 } } \\
    &= [\frac{\tilde{c}_{hard_i}}{\sin(\beta_i)} - \frac{\tilde{c}_{y_i}}{\sin(\alpha_i)}] \cdot \delta(L_i>0)
    \end{aligned}
    \label{equation_backward}
\end{equation}

Turn to the centers, for stability, the update for centers is not strictly using the gradient derived by Eq. \ref{equation_atcl}.
Instead, we use an ``average version" as mentioned in center loss. For simplicity,
we denote ${g}_{1i} = \frac{\tilde{f}_{i}}{\sin(\beta_i)}$ and ${g}_{2i} = \frac{\tilde{f}_{i}}{\sin(\alpha_i)}$,  
then:
\begin{equation}
    \begin{aligned}
    \Delta c_j
    &= \frac{\sum^{M}_{i=0}{g}_{1i} \cdot \delta(L_i>0) \cdot \delta(hard_i=y_j)}{1+\sum^{M}_{i=0}\delta(L_i>0) \cdot \delta(hard_i=y_j)} \\
    & \quad - \frac{\sum^{M}_{i=0}{g}_{2i} \cdot \delta(L_i>0) \cdot \delta(i=y_j)}{1+\sum^{M}_{i=0}\delta(L_i>0) \cdot \delta(i=y_j)}
    \end{aligned}
\end{equation}

\noindent \textbf{Training with softmax loss.}
As a general metric loss, ATCL can be used independently to guide the feature learning and it can converge stably. 
Moreover, we notice that the combination of ATCL and softmax loss functions can further improve the discriminative power of features. 
This is because that softmax loss drops more stably and quickly at the beginning of the training process compared to ATCL. 
And this is useful for centers' training since the initial feature distribution changes a lot during the training process with single ATCL supervision.
Formally, the combination can be written as $L_{total}$:
\begin{equation}
L_{total} = L_{softmax} + \lambda \cdot L_{ATC}
\end{equation}
where $\lambda$ is a trade-off hyper-parameter.  We will discuss the influence of $\lambda$ in the experiment section. 

\subsection{Discussion}
\textbf{Comparison with TCL.}
The main difference between two loss functions is that TCL uses Euclidean distance in the loss design while ATCL uses angular distance.
The angular margin is more suitable for the retrieval task. And it has a more clear geometric interpretation than squared Euclidean distance. 
In fact, in the case of Euclidean space, 
the value of margin depends largely on the magnitude of the elements in the feature vector which makes margin design difficult. 
In contrast, the angular between two vectors is always in $[0,\pi]$ so that the design of angular margin is easier.

\noindent \textbf{Angular margin.}
An appropriate angular margin $\theta$ (between $\frac{\pi}{10}$ and $\frac{\pi}{3}$) can focus on the hard samples in dataset.
If $\theta$ is too large, the loss will concern about the samples that are not important in the retrieval.
We can give an explanation from Eq. \ref{equation_atcl2} and \ref{equation_backward}. 
With a small angular margin $\theta$, 
the losses will not back-propagate for easy samples since their angular distributions are ``perfect" that make $L_i=0$.
As the angular margin grows, the task becomes more difficult and more losses will be propagated. 
As a result, the loss cannot capture the hard samples to which the training should pay more attention.
In our experiment, $\theta=0.7 (\approx 40^\circ)$ has the best performance.

\begin{table}
\begin{center}
\scalebox{0.86}[1]{
\begin{tabular}{lccccc}
\hline
\multirow{2}{*}{Methods} & \multicolumn{2}{c}{ModelNet40} & &\multicolumn{2}{c}{ModelNet10}\\
\cline{2-3} \cline{5-6} & AUC & MAP && AUC & MAP \cr
\hline

ShapeNets & 49.94\% 	& 49.23\% 		&& 69.28\% & 68.26\%\\

DeepPano & 77.63\% 		& 76.81\% 		&& 85.45\% & 84.18\%\\

MVCNN 	& - 				& 78.90\% 		&& - & -\\

GIFT 		& 83.10\% 		& 81.94\% 		&& 92.35\% & 91.12\%\\

GVCNN 	& - 				& 85.70\% 		&& - & -\\
TCL(VGG\_M)&85.59\% 	& 84.35\% 		&& - & -\\
\hline
ATCL 		& 86.52\% 		& 85.35\% 		&& 91.19\% & 90.60\%\\
ATCL+softmax & \textbf{87.23\%} & \textbf{86.11\%} && \textbf{92.49\%} & \textbf{92.07\%}\\
\hline
\end{tabular}}
\end{center}
\caption{The performance comparison with state-of-the-art on ModelNet.}
\label{tab_m40}
\end{table}
\section{Experiment}
In this section, we evaluate the performance of ATCL on two representative large-scale 3D shape datasets and also compare the results with the state-of-art methods. Following we provide comprehensive experiments to discuss the comparison with other metric loss. And we also investigate the influence of parameters, including the angular margin and the hyper-parameter $\lambda$, on the performance of 3D shape retrieval in the last part.

\begin{table*}
\centering

\begin{tabular}{lccccccccc}
\hline
\multirow{2}{*}{Methods} & \multicolumn{3}{c}{Micro} & \multicolumn{3}{c}{Macro}& \multicolumn{3}{c}{Micro + Macro}\\
\cmidrule(lr){2-4}  \cmidrule(lr){5-7} \cmidrule(lr){8-10} & F-measure & MAP & NDCG & F-measure & MAP & NDCG & F-measure & MAP & NDCG\\
\hline
Wang 		& 0.246 & 0.600 & 0.776 & 0.163 & 0.478 & 0.695 					& 0.205 		& 0.539 	& 0.753\\
Li 			& 0.534 & 0.749 & 0.865 & 0.182 & 0.579 & 0.767 						& 0.358 		& 0.664 	& 0.816\\
Kd-network 		& 0.451 	& 0.617 	& 0.814 	& 0.241 & 0.484 & 0.726		& 0.346 		& 0.551 	& 0.770\\
MVCNN 	& 0.612		& 0.734 & 0.843 & 0.416 & 0.662 & 0.793 				& 0.514 		& 0.698 	& 0.818\\
GIFT 		& 0.661 	& 0.811 	& \textbf{0.889} 	& \textbf{0.423} & 0.730 & 0.843		& \textbf{0.542} 	& 0.771 & 0.866\\
\hline
Our & \textbf{0.665} & \textbf{0.843} & 0.885 	& 0.404 & \textbf{0.777} & \textbf{0.851} 		& 0.535 & \textbf{0.810} & \textbf{0.868}\\
\hline
\end{tabular}
\caption{The performance comparison on SHREC16 perturbed dataset}
\label{tab_shrec}
\end{table*}

\noindent\textbf{Implementation.} 
Our experiments are conducted on a server with 2 Nvidia GTX1080Ti GPUs, an Intel Xeon CPU and 128G RAM.
The proposed method is implemented in PyTorch.
For the structure of the CNN, we use VGG-M~\cite{chatfield2014return} as the base network in all our experiments. 
VGG-M has 5 convolution layers and 3 fully-connected layers (\emph{fc6-8}), and we place view pooling layer after (\emph{conv5}). 
The network is pre-trained on ImageNet~\cite{deng2009imagenet} and the centers are initialized by a Gaussian distribution of mean value 0 and standard deviation 0.01.
We use the stochastic gradient descent (SGD) algorithm with momentum 2e-4 to optimize the loss.
The batch size for the mini-batch is set to 20. 
The learning rate for the CNN is 1e-4 and is divided by 10 at the epoch 80.
Specially, the learning rate for centers is always 1e-4 in the training process.
The total training epochs are 120.

In the retrieval process, we render 8 views by Phong reflection in different positions to generate depth images. 
The size of each image is 224x224 pixels in our experiment.
Then all these 8 images are put into the CNN together to generate the model feature.
The features extracted for testing are the outputs of the penultimate layer, \emph{i.e. fc7}.
The cosine distance is used for the evaluation.

\subsection{Retrieval on large-scale 3D datasets}

\noindent\textbf{Dataset.} 
To evaluate the performance of the proposed method, we have conducted 3D shape retrieval experiments on the ModelNet dataset~\cite{wu20153d} and ShapeNetCore55 dataset~\cite{chang2015shapenet}. Next, we will give a detailed introduction of these datasets. \emph{1)} ModelNet Dataset: this dataset is a large-sacle 3D CAD model dataset composed of 127,915
3D CAD models from 662 categories. It has two subsets called ModelNet40 and ModelNet10.
ModelNet40 dataset contains 12,311 models from 40 categories and the second one contains 4,899 
models from 10 categories. For our evaluation experiment, we adopt the same method to split training and test set as mentioned in~\cite{wu20153d}, $i.e.$ randomly select 100 unique models per category from the subset, where the first 80 models are used for training and the rest for testing. \emph{2)} ShapeNetCore55: the dataset from SHape REtrieval Contest 2016~\cite{savva2016shrec} is a subset of the full ShapeNet dataset with clean 3D models and manually verified category and alignment annotations.
This dataset contains 51,190 3D shapes from 55 common categories divided into 204 sub-categories.
There are two versions in this dataset. The first is ``normal" dataset in which the shapes are aligned. And the second 
is more challenging ``perturbed" dataset where all shapes are rotated randomly.
We follow the official training and testing split method to conduct our experiment on perturbed version, where the database is split into three parts, 70\% shapes used for training, 10\% shapes for validation data and the rest 20\% for testing.

The evaluation metrics used in this paper include mean average precision (MAP), area under curve (AUC) and F-measure. Refer to~\cite{wu20153d,savva2016shrec} for their detailed definitions. 

\begin{table}
\begin{center}
\begin{tabular}{lcc}
\hline
Loss function &AUC & MAP\\

\hline

softmax loss 			& 79.67\% 		& 78.28\% \\

softmax loss+center loss  	& 82.24\% 		& 80.95\% \\

triplet loss  			& 80.12\%			& 78.90\% \\

TCL(VGG\_M)			&85.59\% 		& 84.35\% \\

ATCL 					& 86.52\% 		& 85.35\% \\
ATCL+softmax loss	& \textbf{87.23\%} 		& \textbf{86.11\%} \\
\hline
\end{tabular}
\end{center}
\caption{The performance comparison with different loss functions on ModelNet40.}
\label{tab_comploss}
\end{table}

\noindent\textbf{Comparison with the state-of-the-arts.} 
In experiments for ModelNet40 and ModelNet10 datasets, we choose 3D ShapeNets~\cite{wu20153d}, DeepPano~\cite{shi2015deeppano}, MVCNN~\cite{su2015multi}, GVCNN~\cite{feng2018gvcnn}, TCL~\cite{he2018triplet} and GIFT~\cite{bai2016gift} methods for comparison. Note that MVCNN, GVCNN and GIFT are all trained only with softmax loss. The experimental results and comparison among different methods are presented in Tab. \ref{tab_m40}. Our method (ATCL+softmax loss) achieves the best retrieval MAP of $86.11\%$ on ModelNet40 and $92.07\%$ on ModelNet10 respectively. We outperform MVCNN by $7.21\%$ in terms of MAP on ModelNet40 dataset. Compared with GIFT, which leverages an off-line re-ranking technique, the combination of ATCL and softmax loss shows its superiority to improve the MAP by $4.17\%$ on ModelNet40. Besides, as the current state-of-the-art view-based method on ModelNet40, GVCNN is trained with the softmax loss on GoogleLenet and an addition post-processing that the low-rank Mahalanobis metric learning is adopted. Compared with it, our method not only gains $0.41\%$ improvement of MAP on ModelNet40, but also has higher efficiency with the end-to-end training fashion. The above results demonstrate that our deep metric learning method is more effective than softmax loss in retrieval task.

It should be noted that our method is built on the basis of triplet-center loss (TCL). To keep fair comparison, we implement TCL on VGG\_M according to the settings mentioned in~\cite{he2018triplet}. The result shows that our method (ATCL) exceed TCL by $1\%$ in MAP which means that the optimization in angular distance is more effective than that in Euclidean distance. 

For the evaluation in ShapeNetCore55 dataset, 
we take three types of results including macro, micro and mean of macro and micro.
Macro is mainly for providing an unweighted average on the entire database and 
Micro aims to dispose the influence of different model categories sizes. 
In this dataset, two labels are provided and an evaluation called the normalized discounted cumulative gain (NDCG) is defined in this competition. 

We choose Wang~\cite{savva2016shrec}, Li~\cite{savva2016shrec}, K-d network~\cite{klokov2017escape}, MVCNN~\cite{su2015multi} and GIFT~\cite{bai2016gift} methods for comparison.
As shown in Tab. \ref{tab_shrec}, our method achieves the state-of-the-art result. 
Compared with GIFT which uses the re-ranking technique, the F-measure in terms of macro in our method  is lower. 
However, our method is more efficient as no post-processing is used.

\subsection{Compared with other loss functions.} 
The experiments results are shown in Tab. \ref{tab_comploss}.
As a well-designed deep metric loss, ATCL dramatically improves the performance by $7.07\%$ in MAP from single softmax loss supervision.
And the combination with softmax loss, compared to single ATCL, gives an improvement of $0.76\%$ in MAP.
This is owing to the effect of grouping features brought by softmax loss at the beginning of the training process which encourages the centers to converge better.
By using angular metric loss function, our method achieves $0.71\%$ gains in AUC and $1\%$ in MAP compared with TCL.
This demonstrates optimization on angular margin works better than that on Euclidean margin.

\begin{figure}
\centering
\includegraphics[width=1.0\linewidth]{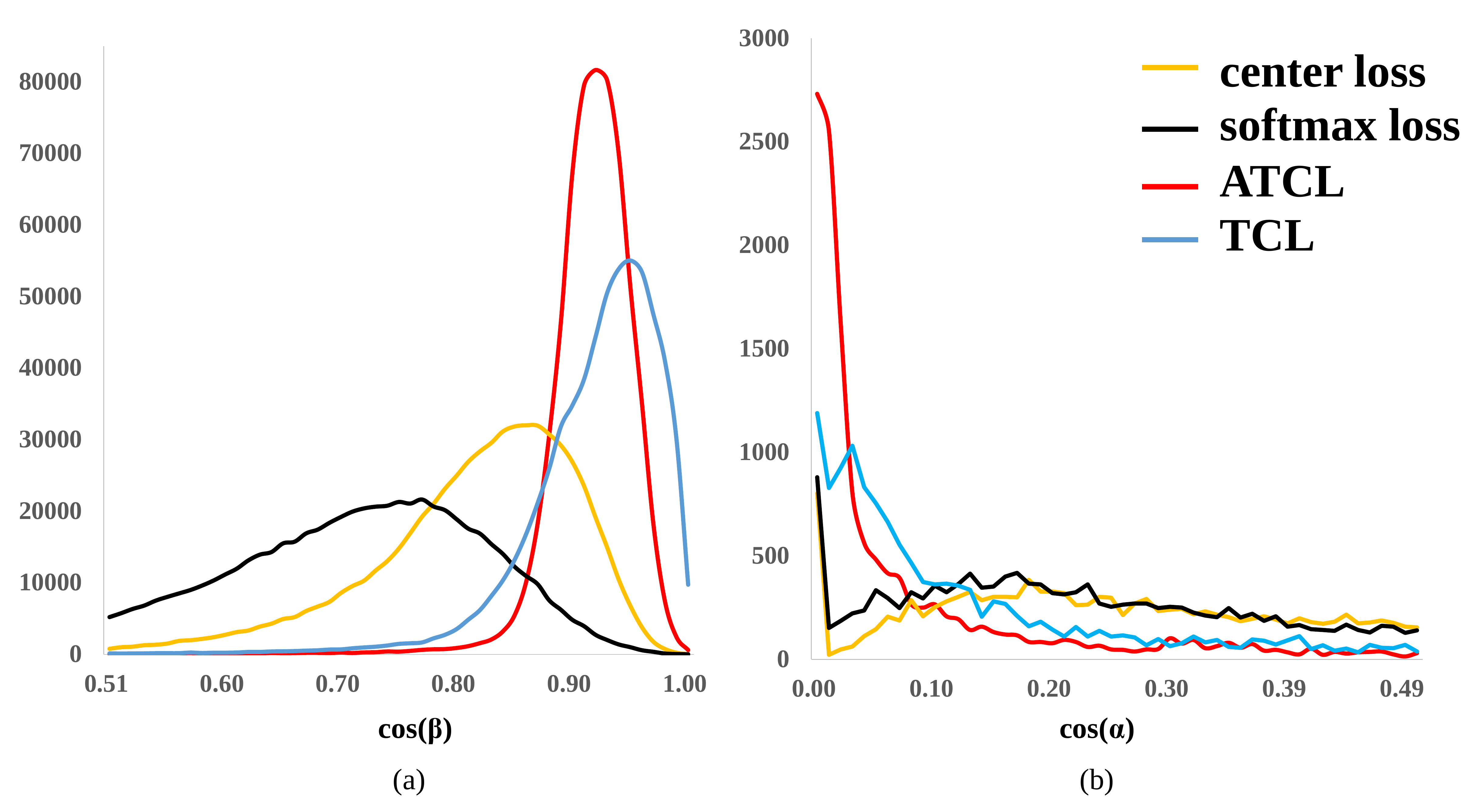}
\caption{Histogram statistics of cosine distance with differrent methods in ModelNet40. 
(a) shows the inter-class cosine distance distribution and (b) is the intra-class cosine distance distribution.
Compared to the Euclidean distance based method, ATCL is more suitable for optimizing cosine distance between features.}
\label{fig_cosdist}
\end{figure}

We show in Fig. \ref{fig_cosdist} the histogram of the cosine distance between features. 
The inter-class distance is shown in Fig. \ref{fig_cosdist} (a) and Fig. \ref{fig_cosdist} (b) is the intra-class distance.
As we can see, with the supervision of ATCL, the distribution of cosine distance is much better than other Euclidean distance based methods.
Most of the intra-class distances in ATCL are nearly 0 while inter-class cosine distances are mainly concentrated between 0.9 and 1.
It convincingly demonstrates that our loss design can significantly improve the discriminative power of embedded features.

\begin{figure}
\centering
\includegraphics[width=1.0\linewidth]{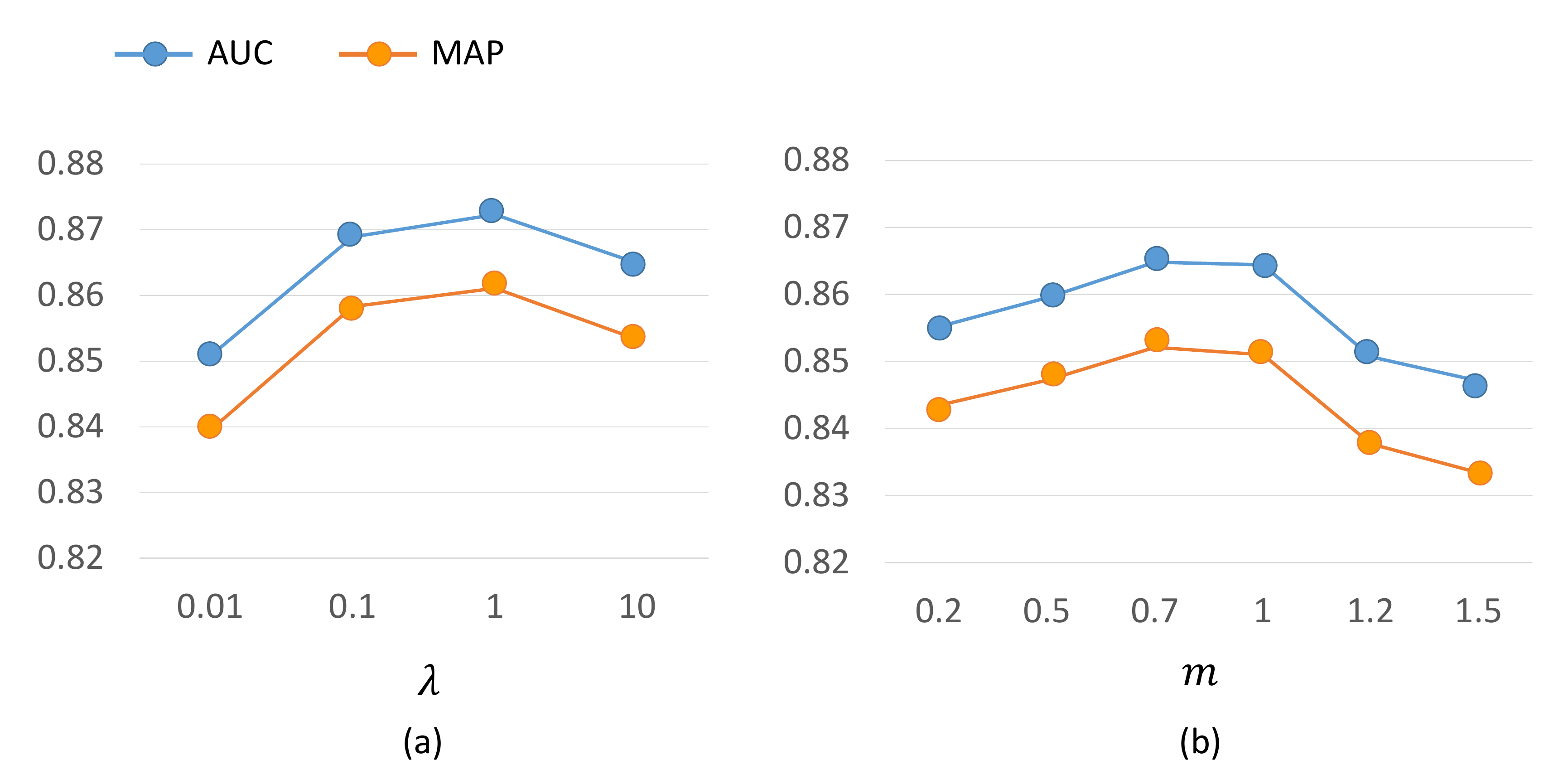}
\caption{The performances of different hyper-parameter settings in ModelNet40.}
\label{fig_parameter}
\end{figure}

\subsection{Parameter analysis}
\label{sec_paramsinfluence}
There two hyper-parameters in our method which are important for the implementation. 
One is the loss weight of ATCL when training jointly with softmax loss $\lambda$ and the other is the angular margin $m$.
We conduct experiments with different values of hyper-parameters to observe the influence on the final performance.

\noindent \textbf{Dicussion on angular margin.}
We use single ATCL as supervision loss and the same parameter settings to conduct experiments in this section.
The results with different $m$ are shown on the left of Fig. \ref{fig_parameter}.
It should be pointed out the margin value here is in radian. 
ATCL converges across a wide range of angular margin from 0.2 ($\approx 11^\circ$) to 1.5 ($\approx 86^\circ$).
After testing different margins, we find that the performance changes slightly when $m$ is not too large.
However, MAP score drops a lot when $m$ becomes larger than 1.
In fact, the angle values between centers are distributed around $\frac{\pi}{2}$.
Although a large angular margin encourages the inter-class distance to become larger, 
it makes every sample in a mini-batch become hard samples (the target is too difficult).
Therefore, a large angular margin has a bad influence on focusing on real hard samples.
And adding too much margin ($ >1.5$) penalty will cause the training divergence. 
The best margin value for the ModelNet is $0.7$ with $86.48\%$ in AUC and $85.21\%$ in MAP.

\noindent \textbf{Dicussion on loss weight.}
We choose $m=0.7$ to conduct all experiments in this section. 
The hyper-parameter $\lambda$ indicates the weight of two tasks.
A small value of $\lambda$ means more attention to the classification task.
For obtaining the best performance using different loss weight, 
we adjust the learning rate in different cases to conduct the experiments.
As $\lambda$ increases from 0.01, the performance first rises and then drops.
The performances are shown on the right of Fig. \ref{fig_parameter}.
We can observe that the best setting is 1:1 which tells us that the combination of two supervisions improves each other.

\section{Conclusion}
In this paper, we propose a novel metric loss function named angular triplet-center loss for the 3D shape retrieval. The loss design inherits the advantage of triplet-center loss and is more powerful than it. Compared with other retrieval methods
based on the metric loss, our loss function focus on optimizing the feature distribution on angle space, providing more explicit discriminative constraints on the shape embedding than Euclidean metric. Extensive results
show that the proposed loss outperforms state-of-the-art methods on two large-scale 3D shape datasets. The learned shape representation is highly discriminative and robust, which is more suitable for 3D shape retrieval task.

\section{Acknowledgements} 
This work is supported by the Science Foundation of Shenzhen City in China (JCYJ20170307100750340), the National Natural Science Foundation of China (No. 61472023) and the Beijing Municipal Natural Science Foundation (No. 4182034).

\bibliographystyle{aaai}
\bibliography{aaai19}

\end{document}